\documentclass[sigconf]{acmart}

\AtBeginDocument{%
  \providecommand\BibTeX{{%
    \normalfont B\kern-0.5em{\scshape i\kern-0.25em b}\kern-0.8em\TeX}}}

\setcopyright{acmcopyright}
\copyrightyear{2022}
\acmYear{2022}
\acmDOI{10.1145/1122445.1122456}
\acmConference[UIST '21 Student Innovation Contest]{UIST '21: ACM Symposium on User Interface Software and Technology}{Oct 10--13, 2021}{Virtual}
\acmBooktitle{UIST '21 Student Innovation Contest: ACM Symposium on User Interface Software and Technology, Oct 10--13, 2021, Virtual}
\acmPrice{15.00}
\acmISBN{978-1-4503-XXXX-X/18/06}

\usepackage{comment}
\begin{document}

\title{
Swarm Fabrication: Reconfigurable 3D Printers and Drawing Plotters Made of Swarm Robots
}

\author{Samin Farajian}
\affiliation{
  \institution{University of Calgary}
  \city{Department of Computer Science}
  \country{}}
\email{farajian.samin@ucalgary.ca}

\author{Hiroki Kaimoto}
\affiliation{
  \institution{The University of Tokyo}
  \city{Interdisciplinary Information Studies}
  \country{}}
\email{hkaimoto@xlab.iii.u-tokyo.ac.jp}

\author{Ryo Suzuki}
\affiliation{
  \institution{University of Calgary}
  \city{Department of Computer Science}
  \country{}}
\email{ryo.suzuki @ucalgary.ca}

\renewcommand{\shortauthors}{}

\begin{abstract}
We introduce Swarm Fabrication, a novel concept of creating on-demand, scalable, and reconfigurable fabrication machines made of swarm robots.
We present ways to construct an element of fabrication machines, such as motors, elevator, table, feeder, and extruder, by leveraging toio robots and 3D printed attachments.
By combining these elements, we demonstrate constructing a X-Y-Z plotter with multiple toio robots, which can be used for drawing plotters and 3D printers.
We also show the possibility to extend our idea to more general-purpose fabrication machines, which include 3D printers, CNC machining, foam cutters, line drawing devices, pick and place machines, 3D scanning, etc.
Through this, we draw a future vision, where the swarm robots can construct a scalable and reconfigurable fabrication machines on-demand, which can be deployed anywhere the user wishes. 
We believe this fabrication technique will become a means of interactive and highly flexible fabrication in the future.
\end{abstract}

\ccsdesc[500]{Human-centered computing~Human-computer interaction}

\keywords{}


\maketitle

\section{Introduction}
Today's digital fabrication machines are not very flexible. 
For example, the users cannot easily bring 3D printers with them and print an object wherever they want. 
In addition, the size and functionality is mostly static---it is often difficult to change the size of printers or modify it from 3D printers to laser cutters. 
The current limitations of these fabrication machines stem from the {\it fixed and inflexible form factors} of these devices---the users cannot easily construct or decompose each element of the machines such as motors, extruders, and gears. 
Therefore, the current fabrication machines are limited in portability, deployability, scalability, and reconfigurability. 

\begin{figure}[htpb]
  \centering
  \includegraphics[width=0.95\columnwidth]{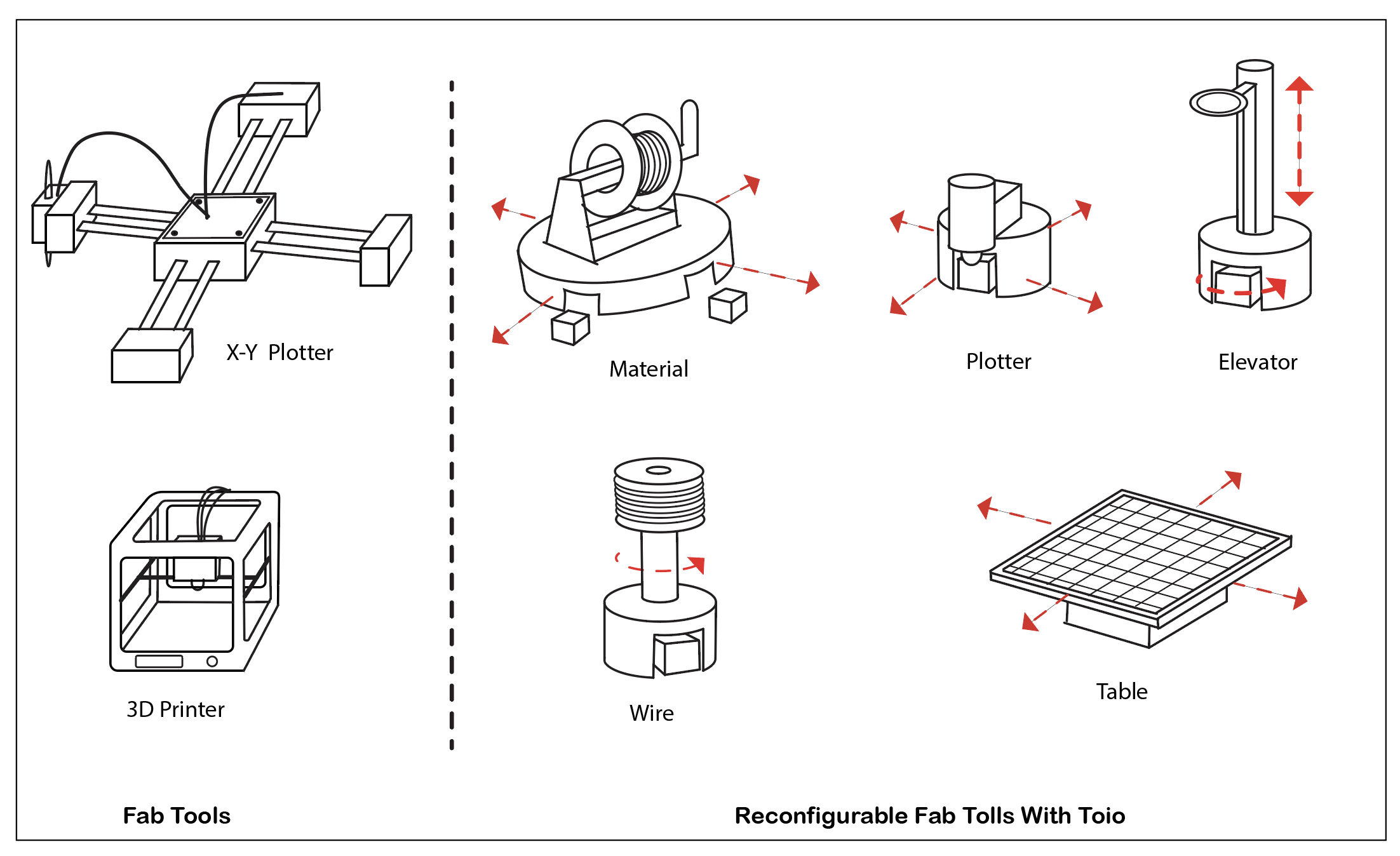}
  \caption{Concept of Swarm Fabricaiton}
  \label{fig:concept}
\end{figure}

In this paper, we propose an alternative approach which we call {\it Swarm Fabrication}, a new concept of dynamically constructing on-demand fabrication machines by leveraging a swarm of robots.
In swarm fabrication, each fabrication machine will be made of swarm robots and custom attachments.
Taking inspiration from HERMITS \cite{nakagaki2020hermits}, we explore an idea where we replace the moving parts and actuators with a small mobile robot.
For example, by combining multiple robots with a custom lead screw, we demonstrate these robots can become a X-Y-Z plotter, whose extruder can move in a 3D position in the programmable fashion (Figure \ref{fig:concept}). 
We envision that this approach enables the fabrication machines more portable, deployable, scalable, and reconfigurable.



\section{Related Work}
\subsubsection*{Modular Fabrication Machines}
Common fabrication machines are usually static and non-reconfigurable. 
To address this limitation, HCI researchers have explored modular fabrication devices. 
For example, Peek et al.~\cite{peek2017cardboard} introduces cardboard machine kit, which enables the user to rapidly prototype the rapid prototyping machines with modular components. 
Similarly, Fabricatable machines~\cite{fossdal2020fabricatable} also explores the software and hardware toolkit to enable this approach. 
Taking inspiration from these works, we further pushes the boundary of modular fabrication by leveraging swarm robots, which enables more portable, scalable, and reconfigurable approach to build on-demand fabrication tools.

\subsubsection*{Small Robots as Fabrication Machines}
There are also several works that investigates the fabrication with small robots. For example, Fiberbots construct an architecture-scale object with a small robot \cite{Kayser2018FIBERBOTS}. 
Koala3D also demonstrates the similar approach for vertical construction \cite{koala3D}. 
Swarm 3D printer \cite{swarm3Dprinting} is also a
Termite Robots \cite{termiterobots} also demonstrates the construction of a large object by collective construction of robots.
The benefits of this approach is to enable the user to fabricate a larger object than the fabrication machines. 
Our work extends this line of work towards the reconfigurable, general-purpose fabrication machines made of swarm robots.

\subsubsection*{Swarm Robots as User Interfaces}
In the field of Swarm Robotics, there are many examples of tangible display representations by swarms of robots \cite{le2016zooids} and display representation by robots with extended functions \cite{suzuki2019shapebots, suzuki2021hapticbots}. 
Especially, our concept is inspired by HERMITS~\cite{nakagaki2020hermits}, which demonstrates to augment the functionality of robots by using mechanical add-ons.
The goal of this work is to show that such mechanical add-ons can be used not only to extend the tangible interactions, but also to create an on-demand prototyping machines. 

\section{System and Implementation}

We propose reconfigurable fabrication tools, which leverage Toio as a functional part of the fabrication machine. Using mechanical attachments, users can easily decompose the fabrication machine, which makes the fabrication tool notably scalable and portable.


\subsubsection*{Toio as Actuator}
Toios could be programmed using P5.Js( p5 is a JavaScript implementation of the popular library called Processing which is based on the JavaScript programming language) to control their movements and leverage them as a functional part of the fabrication tools.
By leveraging this capability, we use Toio as an universal actuator that can actuate mechanical 3D printed attachment, such as lead screws or extruder.


\subsubsection*{X-Y plotter}
\begin{figure}[htpb]
  \centering
  \includegraphics[width=0.95\linewidth]{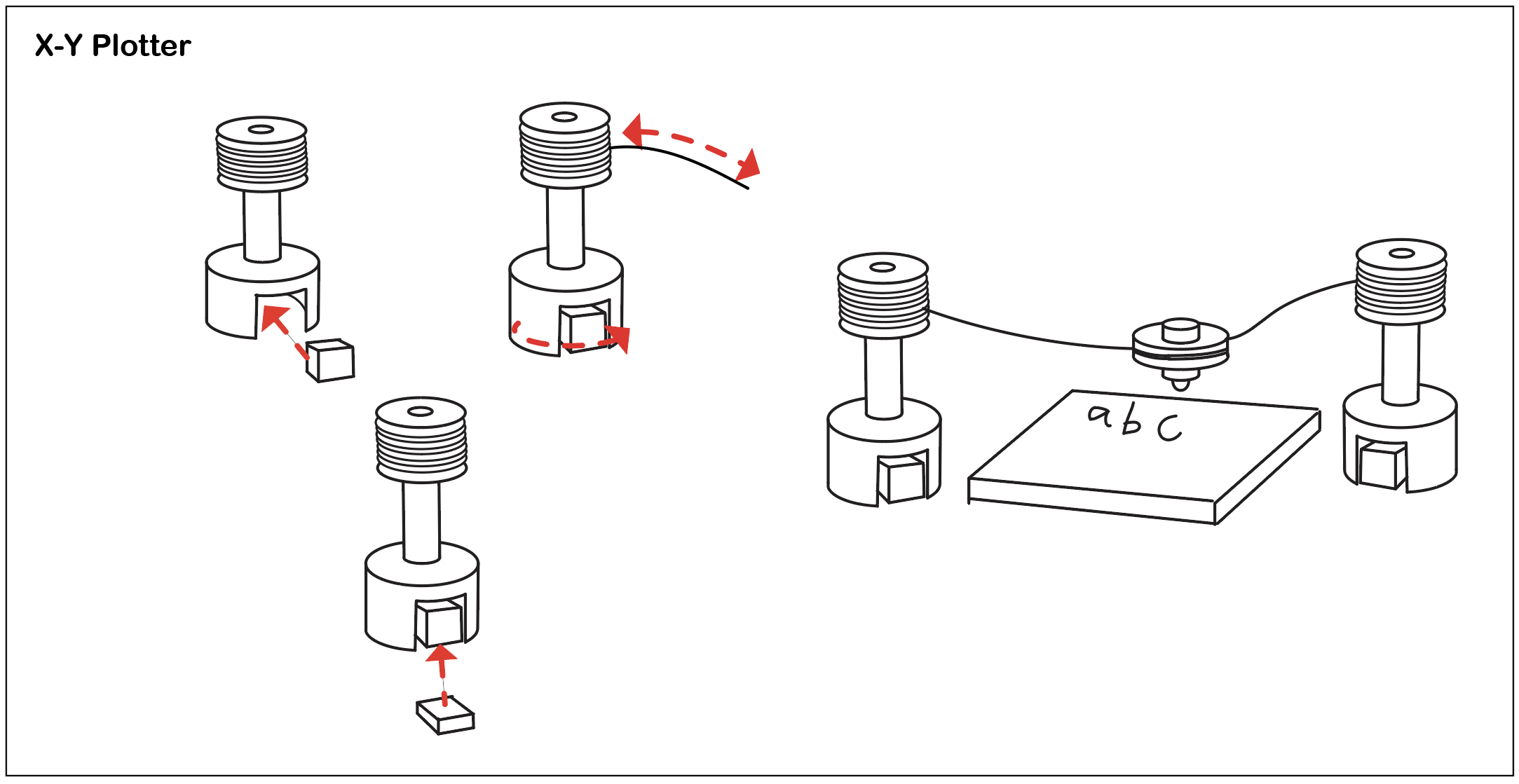}
  \caption{System and Implementation/ x-y Plotter}
  \label{fig:XYplotter}
\end{figure}

We propose to configure an X-Y plotter with Toio and use this as one of the central elements for fabrication tools. In this part, we could configure the X-Y plotter using three Toio robots and custom attachments--- For example, by using two Toios and bridge attachments, we could configure a bridge that can transport. The Toio with plotter attachment can also move on the bridge we made with Toio. By combine these functional Toios and control their position and behavior of each, multiple functional Toios works on as X-Y plotters. We also propose another approach using wire control attachment. This attachment can adjust wire length by rotating motion of Toio. To control wire length with two Toios, we can adjust the position of the plotter on the wire (Figure \ref{fig:XYplotter}). Toio is also moving on vertical surfaces such as whiteboards by attaching a magnet to the bottom of Toio. Therefore, by combining wire attachment and magnet technic, we also configure X-Y plotter on vertical surfaces similar to Scribit\cite{scribit}. 

\begin{figure}[htpb]
  \centering
  \includegraphics[width=0.95\linewidth]{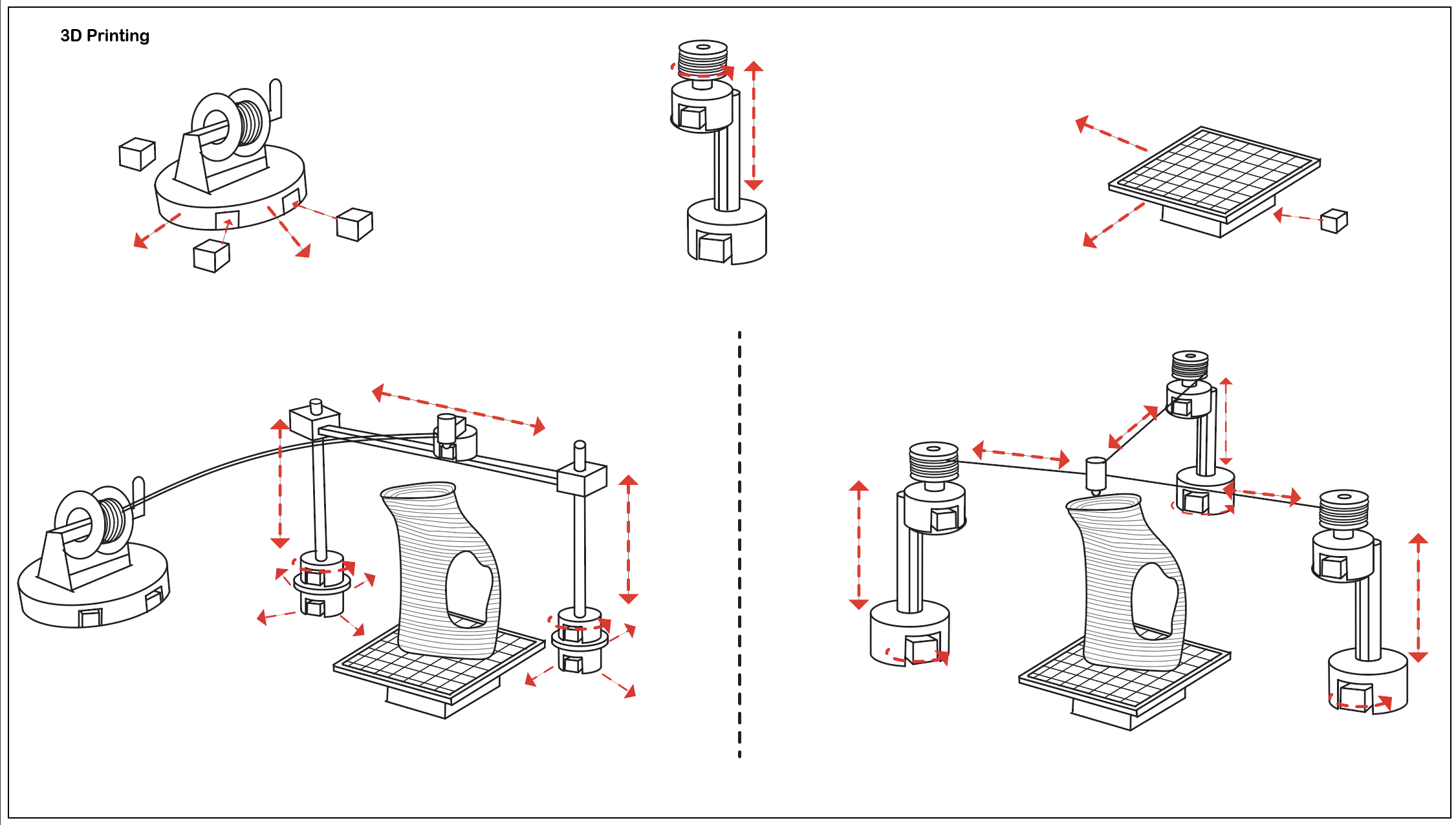}
  \caption{System and Implementation/ 3D Printing }
  \label{fig:3Dprinting}
\end{figure}

\subsubsection*{3D printing}
Using mechanical attachments used on the X-Y plotter and other mechanical attachments, we also propose to configure the 3D printing function with Toio. We could also utilize bridge tools using Toios for 3D printing with other attachments: material extruder attachment, modeling table attachment, and material container attachment. The Toio with extruder attachment also moves on the bridges made with two Toios, and the Toio with the table attachment also moves. These two functional Toios will move to print 3D objects based on the g-code(Figure \ref{fig:3Dprinting}). We also propose a wire control system as well as the X-Y plotter part. In this case, we could adjust the position of the extruder with 3 wires based on Toio behavior (Figure \ref{fig:3Dprinting}). We consider these implementations also be applicable to other fabrication machines such as CNC machining, foam cutters, line drawing devices, pick and place machines, 3D scanning, etc.

\section{Future Work}
In this demo, we specifically focused on the basic elements---X-Y plotter, wire drawing plotter, and 3D printer. 
However, the concept of swarm fabrication can be extended to more general-purpose fabrication machines. 
For example, by leveraging these reconfigurable elements, we could also construct CNC machining, foam cutters, line drawing devices, pick and place machines, 3D scanning, etc.
The autonomous and reconfigurable design opens up a new opportunity for fabrication machines. 
For example, imagine that the extruder can be automatically swapped with different robots to reconfigure from a 3D printer to multi-material 3D printer, pick-and-place machine, robotic assembly machines, or drawing plotters. 
With this, we envision that it is possible to make 3D printers as a reconfigurable robotic assembly device, similar to~\cite{katakura20193d}.
We also envision that this enables the scalable mobile fabrication~\cite{roumen2016mobile, peek2017popfab}, where the user can easily construct the fabrication machine on demond.


\balance
\bibliographystyle{ACM-Reference-Format}
\bibliography{references}

\end{document}